\documentclass{z-image}
\usepackage{natbib}
\usepackage{multirow}
\usepackage{makecell}
\usepackage{etoolbox}

\definecolor{scholarblue}{rgb}{0.21,0.49,0.74}
\definecolor{bluelink}{RGB}{0,113,188}
\hypersetup{
    colorlinks=true,
    citecolor=scholarblue,
    linkcolor=red!93!black,
    urlcolor=bluelink,
    pdftitle={FastVR: Efficient Streaming Video Restoration with One-Step Diffusion},
    pdfauthor={Xiaoxu Chen, Qin Yang, Haoran Bai, Sibin Deng, Ying Chen}
}

\titlespacing*{\paragraph}{0pt}{0.25em}{1em}

\fancypagestyle{fastvrfirst}[firststyle]{%
    \fancyhead[L]{\includegraphics[height=30pt]{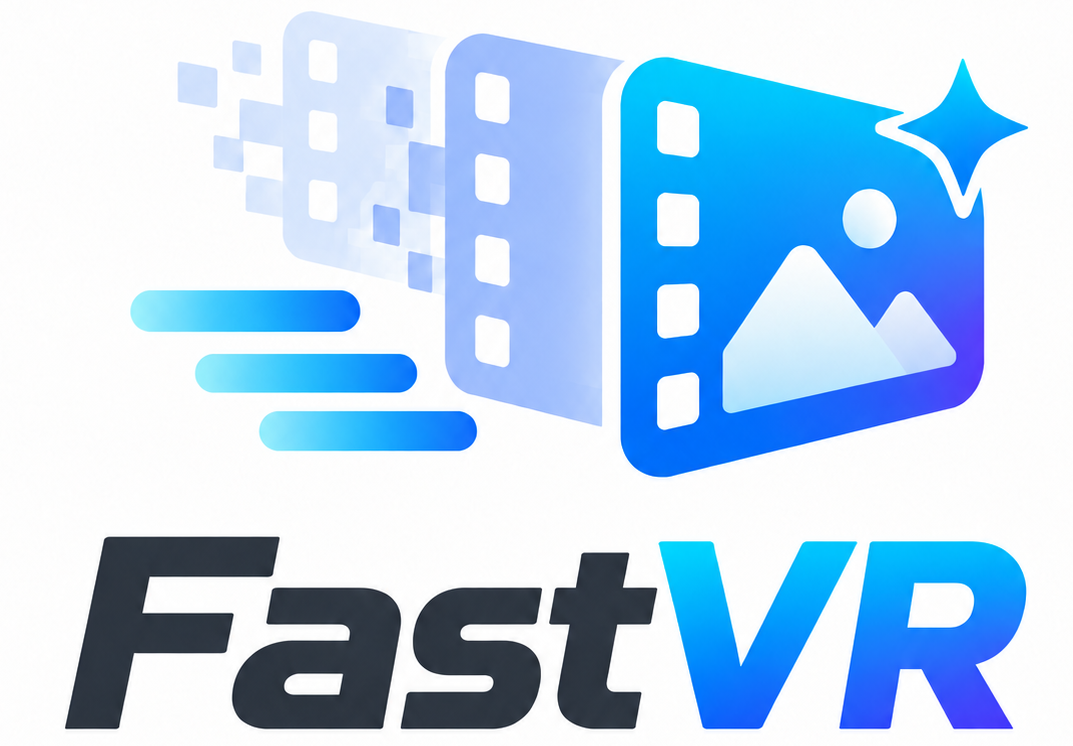}}%
    \fancyfoot[L]{\normalfont\footnotesize
        \textsuperscript{*} Equal contribution.\quad
        \textsuperscript{$\dagger$} Corresponding author.}%
}
\renewcommand\Affilfont{\normalfont\fontsize{11}{15}\selectfont\centering}

\title{\centering FastVR: Efficient Streaming Video Restoration\\with One-Step Diffusion}
\author{%
    Xiaoxu Chen\textsuperscript{1,*}\quad
    Qin Yang\textsuperscript{1,2,*}\quad
    Haoran Bai\textsuperscript{1}\quad
    Sibin Deng\textsuperscript{1}\quad
    Ying Chen\textsuperscript{1,$\dagger$}\\
    \textsuperscript{1}Alibaba Group\qquad
    \textsuperscript{2}Xidian University
}

\newcommand{\reportlinks}{%
    \begingroup
        \centering\small
        \renewcommand{\arraystretch}{1.4}
        \begin{tabular}{@{}c@{\hspace{0.5em}}l@{\hspace{0.8em}}l@{}}
            \raisebox{-1.5pt}{\includegraphics[height=1.05em]{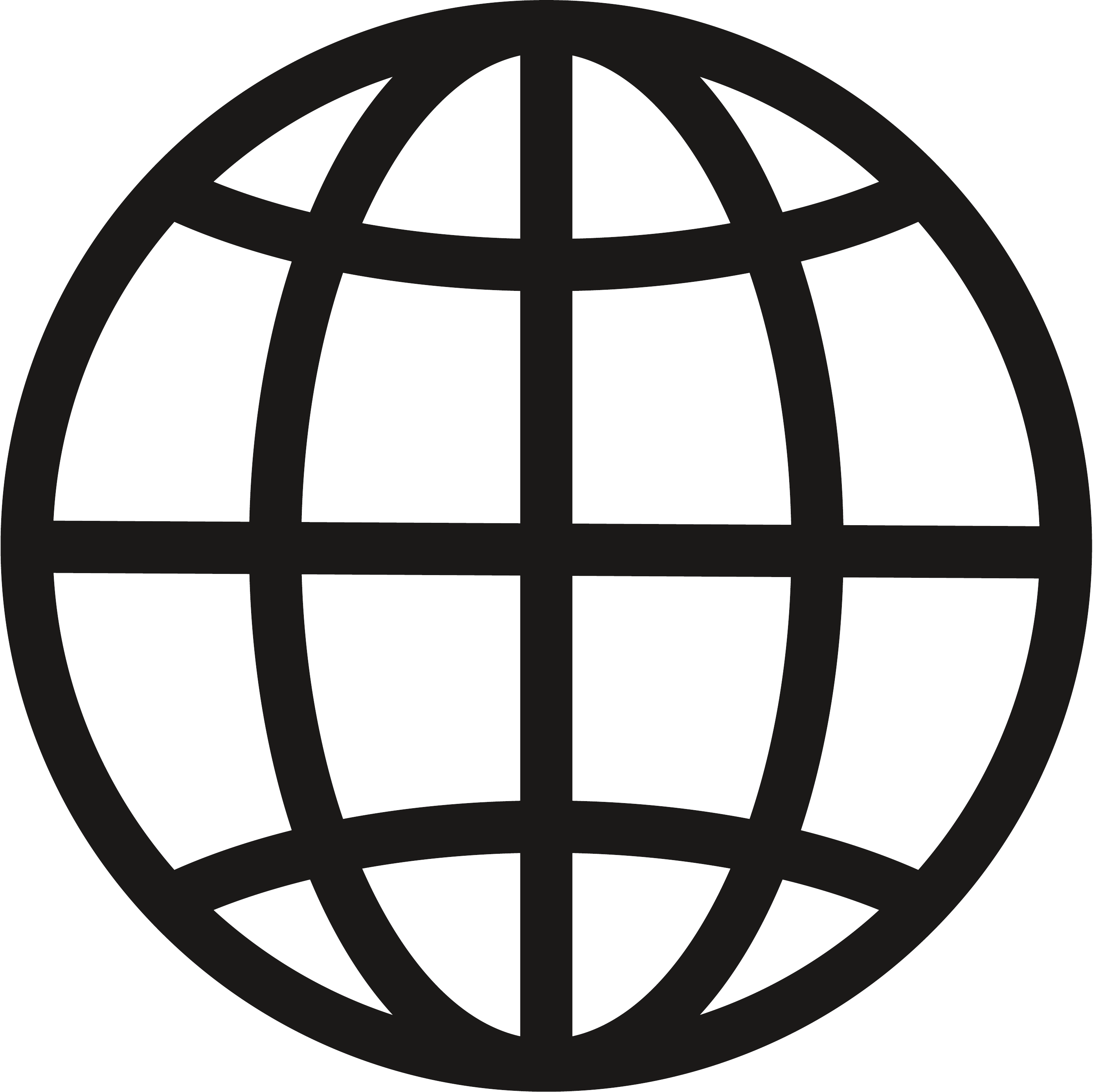}} & \textbf{Project Page} & \url{https://chenxx89.github.io/projects/fastvr/} \\
            \raisebox{-1.5pt}{\includegraphics[height=1.05em]{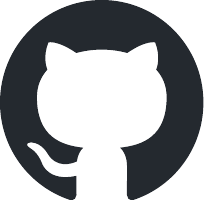}} & \textbf{GitHub} & \url{https://github.com/chenxx89/FastVR} \\
            \raisebox{-1.5pt}{\includegraphics[height=1.05em]{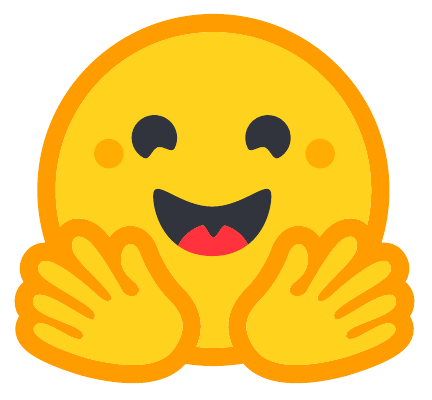}} & \textbf{Hugging Face} & \url{https://huggingface.co/chenxx89/FastVR}
        \end{tabular}\par
    \endgroup
    \vspace{1ex}
}

\newcommand{\reportteaser}{%
    \begin{minipage}{\linewidth}
        \centering
        \captionsetup{hypcap=false}
        \makebox[\linewidth][c]{%
            \includegraphics[width=1.05\linewidth, pagebox=cropbox]{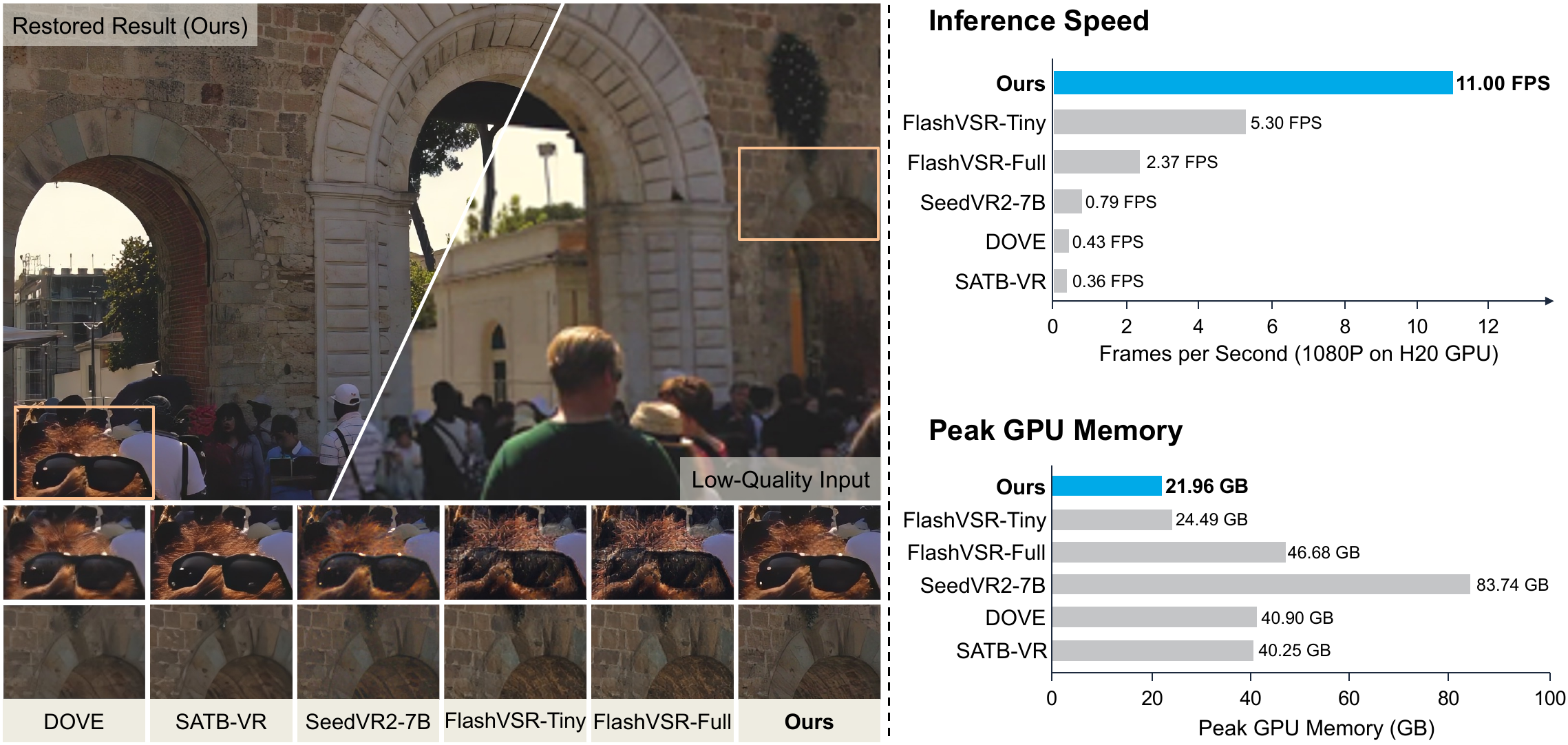}%
        }
        \captionof{figure}{Visual quality and inference efficiency of FastVR. Left: A restoration example and zoomed-in comparisons with existing methods. Right: Inference speed and peak GPU memory usage when processing a 120-frame 1080p video on a single NVIDIA H20 GPU.}
        \label{fig:teaser}
    \end{minipage}
    \par\vspace{1.5ex}
}
\patchcmd{\maketitle}{\vskip20pt}{\vskip12pt}{}{
    \PackageError{FastVR}{Cannot adjust title spacing}{Check z-image.cls.}
}
\patchcmd{\maketitle}{\vskip30pt}{\vskip4pt}{}{
    \PackageError{FastVR}{Cannot adjust teaser spacing}{Check z-image.cls.}
}
\pretocmd{\abscontent}{\reportlinks\reportteaser}{}{
    \PackageError{FastVR}{Cannot insert the links and teaser}{Check z-image.cls.}
}

\begin{abstract}
Diffusion-based video restoration recovers realistic details, but its practical deployment is limited by two efficiency bottlenecks: costly VAE encoding and decoding, and the quadratic cost of full self-attention in diffusion transformers (DiTs).
This paper presents FastVR, a streaming video restoration framework built on a one-step diffusion model, which delivers strong restoration quality and temporal consistency while processing 1080p video at 11 FPS on a single H20 GPU.
To improve inference efficiency, FastVR combines a lightweight VAE with chunk-wise causal attention, which substantially reduces the computational cost.
During training, it further adopts velocity consistency regularization and continuous trajectory learning, which improve restoration quality.
Extensive experiments show that FastVR is more efficient than the evaluated diffusion baselines while achieving state-of-the-art performance on synthetic and real-world benchmarks.
We hope that this work supports further progress in the community.

\end{abstract}

\begin{document}
\maketitle\par
\thispagestyle{fastvrfirst}

\section{Introduction}
Video restoration aims to recover visually faithful and temporally consistent content from videos degraded by blur, noise, compression, downsampling, and other artifacts. In practical deployment, high-resolution long videos generally need to be processed with low latency under limited computation and memory budgets. Although traditional reconstruction-based methods \cite{basicvsr,basicvsrpp} can achieve high accuracy, they tend to produce overly smooth textures when severe degradation removes high-frequency details. In recent years, diffusion-based video restoration \cite{dove,seedvr2,vivid,flashvsr} has attracted considerable attention, since learned generative priors can restore realistic textures and fine details. Image diffusion priors improve spatial detail, but they do not model temporal dynamics explicitly and can therefore break temporal consistency across frames. Video diffusion models alleviate this limitation by modeling motion and appearance jointly. STAR \cite{star} adapts a text-to-video prior to real-world video super-resolution, improving degradation removal while preserving temporal consistency. Vivid-VR \cite{vivid} further shows that a large text-to-video DiT can generate photorealistic textures without sacrificing temporal consistency. These studies suggest that video diffusion priors provide a strong basis for recovering spatial detail and maintaining temporal consistency, although the gains come at a high computational cost.

The first factor that limits efficiency is the cost of iterative sampling. Conventional methods repeatedly evaluate a large denoising network, so latency grows roughly linearly with the number of sampling steps. Vivid-VR, for example, uses a large video diffusion backbone with 50 inference steps. Recent one-step methods have achieved notable progress. DOVE \cite{dove} fine-tunes a pretrained video diffusion model directly for one-step real-world video super-resolution and achieves a large speedup over multi-step baselines. However, the authors also observe that endpoint regression tends to yield over-smoothed results, which motivates an additional refinement stage in pixel space. SeedVR2 \cite{seedvr2} combines progressive distillation with adversarial post-training in order to preserve restoration capability after compressing a multi-step process into a single step. DUO-VSR \cite{duovsr} strengthens one-step optimization through distribution matching, feature-level adversarial supervision, and preference refinement. Although the one-step methods described above improve perceptual quality, they discard the progressive correction mechanism of diffusion sampling and remain expensive on high-resolution videos.

One-step modeling removes much of this cost, but two other system-level bottlenecks then become dominant. In the DiT, the cost of global self-attention grows quadratically with the number of tokens. Moreover, although latent compression by the VAE reduces the denoising cost, reconstruction through a large causal video VAE can dominate the overall latency. SeedVR2 \cite{seedvr2} reports that the causal video VAE accounts for more than 95\% of the total runtime on a 100-frame 720p video, and DUO-VSR \cite{duovsr} likewise identifies VAE processing as the main overhead in a one-step pipeline. Reducing the number of sampling steps is therefore not sufficient for high-resolution restoration. Recently, several streaming methods have adopted comprehensive solutions to further improve inference efficiency. FlashVSR \cite{flashvsr} introduces causal sparse attention, KV cache, one-step distillation, and a lightweight decoder, reaching near real-time throughput on long videos. The analysis in FlashVSR also shows that VAE decoding becomes a dominant component once denoising is compressed into a single step, while dense attention remains inefficient at high resolution. Such results reshape the central research question: the challenge is no longer only to reduce the number of diffusion steps, but to achieve a better end-to-end balance among autoencoding cost, attention complexity, optimization stability, restoration fidelity, and temporal consistency.

We present FastVR, a one-step diffusion framework for high-quality streaming video restoration that targets the two costs which become dominant after sampling acceleration. At the representation level, FastVR adopts a lightweight VAE to reduce the cost of encoding degraded frames and decoding restored frames. At the denoising level, FastVR combines spatial tiling with chunk-wise causal attention to limit spatial computation and temporal attention context, reducing token interaction costs and supporting online processing without access to future chunks. To compensate for the limited use of progressive correction in a one-step mapping, FastVR constructs a continuous restoration trajectory between low-quality and high-quality videos, so that the model observes intermediate states along the transformation from a degraded video to a clean one rather than a single supervised endpoint. FastVR further imposes an explicit velocity consistency regularizer across sampled trajectory points, which encourages compatible restoration velocities at different sampling positions and constrains the learned vector field, thereby improving the optimization stability of one-step prediction. FastVR runs at 11 frames per second on 1080p video with a single NVIDIA H20 GPU, while maintaining strong restoration quality and temporal consistency. In summary, our main contributions are as follows:
\begin{itemize}
    \item We design an efficient framework for high-resolution streaming video restoration, in which a lightweight VAE and chunk-wise causal attention reduce the encoding, decoding, and attention costs of one-step inference. 
    \item We introduce explicit velocity consistency regularization into the learning of continuous restoration trajectories, which improves both the optimization stability and the detail recovery ability of the one-step model. 
    \item Extensive experiments on synthetic and real-world benchmarks show that FastVR attains competitive restoration quality and temporal consistency while reaching 11 FPS on 1080p video with a single H20 GPU. 
\end{itemize}

Figure~\ref{fig:teaser} illustrates the restoration quality and inference efficiency of FastVR compared with existing methods.

\section{Related Work}
\label{related-work}
\subsection{Diffusion-based Video Restoration.}
Conventional VSR methods trained on synthetic or composite degradations~\cite{basicvsr,basicvsrpp} can still struggle to recover realistic fine details under severe degradation. Diffusion models~\cite{ddpm, score-model} instead bring a strong generative prior to this task. This prior proves highly effective in image restoration~\cite{isr1,isr2,isr3}, where realistic details can be faithfully synthesized. However, extending this success to video is not straightforward, as temporal consistency must be preserved in addition to spatial fidelity. Early attempts therefore attached temporal modules to pretrained image backbones: Upscale-A-Video~\cite{upscale} propagates latents along optical-flow trajectories, MGLD-VSR~\cite{mgld-vsr} steers sampling with motion-aware losses, and DiffVSR~\cite{diffvsr} augments the backbone with multi-scale temporal attention and a staged training scheme. However, the underlying 2D prior limits their robustness under severe spatiotemporal corruption. This limitation has driven a shift toward video-native Diffusion Transformers, whose pretraining on large-scale text-to-video data~\cite{cogvideox,wan} provides a stronger motion prior. Building on this foundation, SeedVR~\cite{seedvr}, STAR~\cite{star}, and Vivid-VR~\cite{vivid} introduce restoration-specific architectures and objectives, and currently set the state of the art in detail fidelity.

\subsection{Video Diffusion Acceleration.}
Despite this fidelity, the iterative denoising of multi-step sampling imposes prohibitive inference latency, especially at high resolutions. To reduce this cost, a line of work condenses the sampling process into a single forward pass via distillation~\cite{dmd,dmd2}, adversarial post-training~\cite{apt}, or rectified-flow techniques~\cite{liu2023flow}, which have recently been extended to video restoration. DOVE~\cite{dove} trains a text-to-video model for direct one-step generation through a two-stage latent-to-pixel scheme, SeedVR2~\cite{seedvr2} reaches a single step via progressive distillation and adversarial post-training, and DUO-VSR~\cite{duovsr} unifies distribution matching with adversarial supervision through dual-stream distillation. Even after the sampling steps are compressed to one, the VAE and the bidirectional attention in the DiT become the dominant latency sources in high-resolution super-resolution. To address this, FlashVSR~\cite{flashvsr} couples one-step distillation with sparse causal attention and a compact decoder, forming the first diffusion-based streaming framework toward real-time VSR.

\section{Method}
\label{method}

\begin{figure}[!t]
    \centering
    \includegraphics[width=\linewidth, pagebox=cropbox]{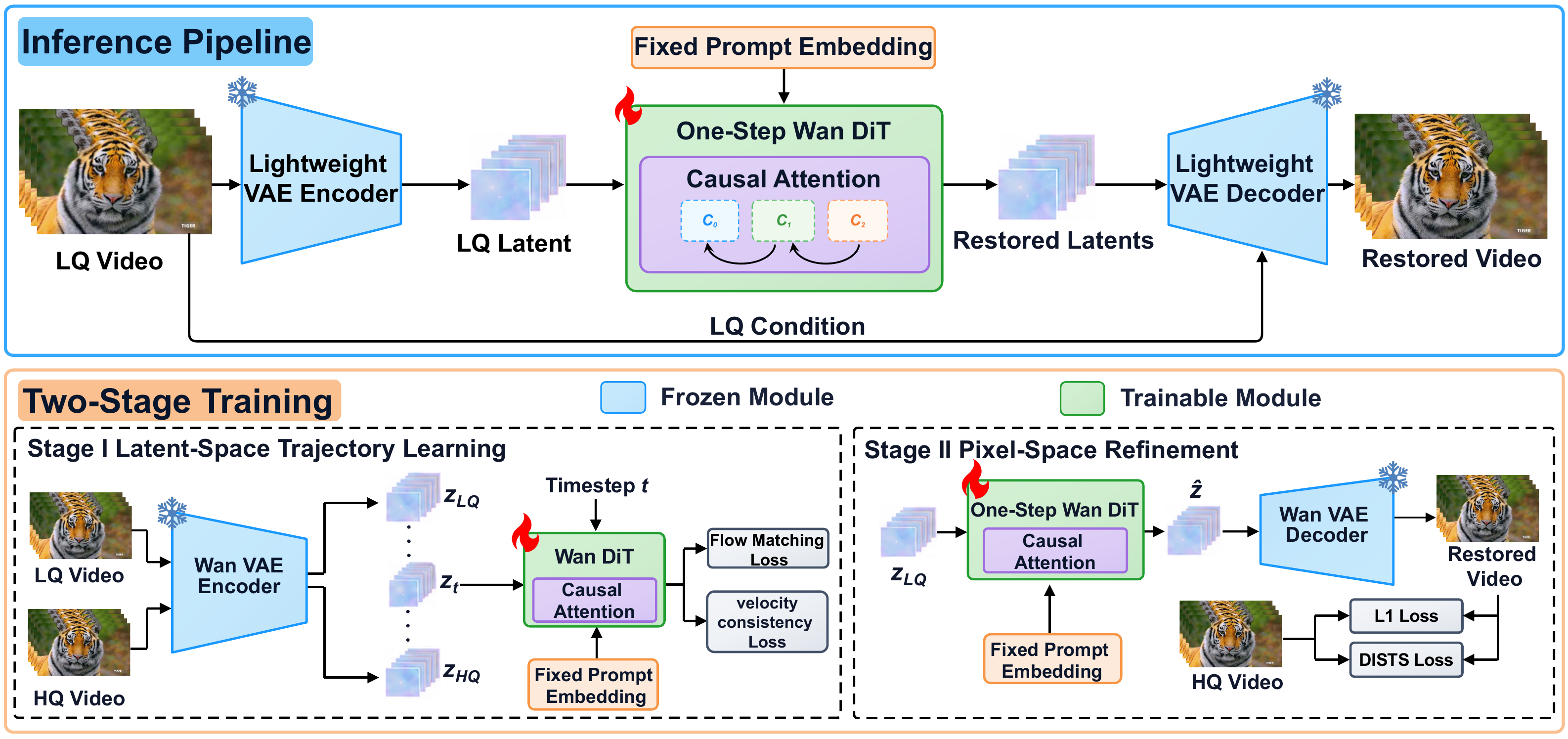}
    \caption{Overview of FastVR. Top: one-step streaming inference with a lightweight VAE and chunk-wise causal attention, with LQ frames conditioning the decoder. Bottom: two-stage DiT training with the frozen Wan VAE, using latent-space trajectory learning and velocity consistency in Stage~I, followed by pixel-space $\ell_1$ and DISTS supervision in Stage~II.}
    \label{fig:overview}
\end{figure}

FastVR is a one-step streaming restoration model built on Wan2.2-TI2V-5B~\cite{wan}. As illustrated in Fig.~\ref{fig:overview}, it combines a lightweight VAE with a chunk-wise causal attention pattern to reduce the two dominant costs of high-resolution video restoration: the cost of the VAE and the quadratic cost of bidirectional attention. During inference, a frozen lightweight encoder first maps a low-quality (LQ) video into a compact latent representation, the DiT predicts a restoration velocity in one forward pass per temporal chunk and spatial tile, and a frozen lightweight decoder reconstructs the restored video from the updated latents and the LQ video. To learn the restoration mapping, the DiT is trained in two stages using the frozen Wan VAE. Stage I learns a continuous restoration trajectory in the latent space and regularizes the predicted velocity along a model-induced trajectory. Stage II then optimizes the model under pixel-space supervision, targeting details that latent-space regression alone does not preserve.

\subsection{Model Architecture}
\label{sec:architecture}

\paragraph{Lightweight VAE.}
To reduce the encoding and decoding overhead during inference, we adopt a lightweight VAE based on lightweight autoencoding~\cite{tae} and conditional decoding~\cite{flashvsr}. Its encoder maps LQ frames into a latent representation compatible with the pretrained Wan DiT. The decoder reconstructs the video from the restored latents while using the LQ frames as an additional condition. This supplies structural information directly from the input and reduces the burden of reconstructing the video solely from compressed latents. We train the lightweight VAE separately from the DiT, using a combination of pixel-wise $\ell_1$ and LPIPS perceptual losses~\cite{zhang2018lpips} between the decoded output and the original HQ video, balancing reconstruction fidelity and perceptual quality. Both encoding and decoding support incremental temporal processing with cached features, allowing bounded intermediate memory during streaming inference. The lightweight VAE is frozen for inference, while the two-stage DiT training uses the original frozen Wan VAE encoder and decoder.

\paragraph{Chunk-wise causal attention.}
The pretrained Wan DiT employs bidirectional self-attention, whose complexity is quadratic in video length. We reduce this cost by partitioning the latent sequence into non-overlapping chunks of $k$ temporal positions and restricting each chunk to attend only to itself and its immediate predecessor. Let $T$ denote the number of latent frames and $i,j\in\{0,\ldots,T-1\}$ the temporal indices of a query and a key, respectively. We define the temporal visibility mask $\mathbf{M}\in\{0,1\}^{T\times T}$ as
\begin{equation}
    M_{i,j} =
    \begin{cases}
        1, &
        \left\lfloor i/k \right\rfloor
        -
        \left\lfloor j/k \right\rfloor
        \in \{0,1\},\\
        0, & \text{otherwise}.
    \end{cases}
    \label{eq:window_causal_mask}
\end{equation}
The mask is applied in every self-attention layer, preserving bidirectional attention within each chunk while limiting temporal context to one preceding chunk.

During training, the entire latent sequence is processed in parallel, with $\mathbf{M}$ enforcing the chunk-wise causal attention masks. During inference, we sequentially restore video chunks using a streaming inference scheme and employ KV caching for efficient computation. For this sequential inference scheme, with fixed chunk size $k$ and spatial resolution, the temporal attention cost scales as $\mathcal{O}(Tk)$, compared with $\mathcal{O}(T^2)$ for bidirectional attention, while the KV-cache size remains $\mathcal{O}(k)$ independent of video length.

\subsection{Two-Stage Latent--Pixel Training}
\label{sec:training}

We optimize FastVR in two stages while keeping all encoder and decoder parameters fixed. The first stage learns the LQ-to-HQ transformation as a continuous vector field in the latent space, whereas the second stage directly optimizes the decoded result produced by one-step prediction. This design first establishes the restoration mapping under efficient latent-space supervision and subsequently adapts the predicted latent to the reconstruction characteristics of the Wan VAE decoder.

Given an HQ video $\mathbf{x}_{\mathrm{HQ}}$, we synthesize the paired LQ observation $\mathbf{x}_{\mathrm{LQ}}$ online using the RealBasicVSR
degradation pipeline~\cite{realbasicvsr}. The LQ and HQ videos are encoded by the frozen Wan VAE encoder~\cite{wan}:
\begin{equation}
    \mathbf{z}_{\mathrm{LQ}}
    =\mathcal{E}_{\mathrm{Wan}}(\mathbf{x}_{\mathrm{LQ}}),
    \qquad
    \mathbf{z}_{\mathrm{HQ}}
    =\mathcal{E}_{\mathrm{Wan}}(\mathbf{x}_{\mathrm{HQ}}).
    \label{eq:latent_endpoints}
\end{equation}

\noindent\textbf{Stage I: Latent-space trajectory learning.}
Unlike endpoint regression~\cite{dove}, we supervise the restoration mapping over a continuous path between $\mathbf{z}_{\mathrm{HQ}}$ and $\mathbf{z}_{\mathrm{LQ}}$. Let $\sigma_t$ denote the noise level associated with timestep $t$. We parameterize the restoration trajectory such that
$t=0$ corresponds to the HQ endpoint $\mathbf{z}_{\mathrm{HQ}}$ with $\sigma_0=0$, whereas $t=\tau$ corresponds to the LQ endpoint $\mathbf{z}_{\mathrm{LQ}}$ with noise level $\sigma_\tau$. The intermediate latent is defined as follows:
\begin{equation}
    \mathbf{z}_t
    =\frac{t}{\tau}\mathbf{z}_{\mathrm{LQ}}
    +\left(1-\frac{t}{\tau}\right)\mathbf{z}_{\mathrm{HQ}}.
    \label{eq:trajectory}
\end{equation}
The corresponding velocity with respect to $\sigma_t$ is
\begin{equation}
    \mathbf{u}_t
    =
    \frac{\partial \mathbf{z}_t}{\partial \sigma_t}
    =
    \frac{\mathbf{z}_{\mathrm{LQ}}-\mathbf{z}_{\mathrm{HQ}}}
         {\sigma_\tau}.
    \label{eq:target_velocity}
\end{equation}
We train the DiT velocity predictor $v_\theta$ using
\begin{equation}
    \mathcal{L}_{\mathrm{traj}}
    =
    \mathbb{E}_{\mathbf{x}_{\mathrm{HQ}},
                t}
    \left[
        \left\|
        v_\theta(\mathbf{z}_t,t,\mathbf{c}_{\varnothing})
        -\mathbf{u}_t
        \right\|_2^2
    \right],
    \label{eq:trajectory_loss}
\end{equation}
where $\mathbf{c}_{\varnothing}$ denotes the empty-text condition. Random timestep sampling exposes the model to latent states with varying degradation levels, enabling it to learn restoration velocities throughout the LQ-to-HQ trajectory rather than only the endpoint mapping.

Although $\mathcal{L}_{\mathrm{traj}}$ provides velocity supervision
at states sampled from the reference trajectory, the latent update
during inference is determined by the model's own prediction. An
inaccurate velocity may therefore displace the predicted latent from
the reference trajectory and directly introduce errors into the
one-step restoration result. To mitigate this discrepancy, we impose
self-distilled velocity consistency along a model-induced trajectory.

Specifically, for a sampled timestep $t$, we draw
$t'\sim\mathcal{U}(0,t)$ and perform an Euler update from
$\sigma_t$ to $\sigma_{t'}$, where $\sigma_{t'}<\sigma_t$:
\begin{equation}
\begin{aligned}
  \mathbf{v}_t
  &=
  v_\theta(
      \mathbf{z}_t,
      t,
      \mathbf{c}_{\varnothing}),\\
  \widetilde{\mathbf{z}}_{t'}
  &=
  \operatorname{sg}\!\left[
      \mathbf{z}_t+
      (\sigma_{t'}-\sigma_t)\mathbf{v}_t
  \right].
\end{aligned}
\label{eq:euler_state}
\end{equation}
Here, $\operatorname{sg}[\cdot]$ denotes the stop-gradient operator.
We then evaluate the velocity at the propagated state
$\widetilde{\mathbf{z}}_{t'}$ and use the detached prediction as the
self-distillation target:
\begin{equation}
  \mathcal{L}_{\mathrm{vc}}
  =
  \mathbb{E}_{(\mathbf{x}_{\mathrm{LQ}},
  \mathbf{x}_{\mathrm{HQ}}),\,t,t'}
  \left[
      \left\|
      \mathbf{v}_t
      -
      \operatorname{sg}\!\left[
          v_\theta(
              \widetilde{\mathbf{z}}_{t'},
              t',
              \mathbf{c}_{\varnothing})
      \right]
      \right\|_2^2
  \right].
  \label{eq:velocity_consistency}
\end{equation}
This objective encourages consistent velocity predictions between a
reference state and the state reached by the model's own update. The
overall objective for Stage~I is
\begin{equation}
  \mathcal{L}_{\mathrm{I}}
  =
  \mathcal{L}_{\mathrm{traj}}
  +
  \mathcal{L}_{\mathrm{vc}}.
  \label{eq:stage_one_loss}
\end{equation}

\noindent\textbf{Stage II: Pixel-space refinement.}
Stage~I provides latent-space supervision but does not directly account for the reconstruction characteristics of the VAE decoder. Because the mapping from the latent space to the pixel space is nonlinear, a small latent-space error does not necessarily imply high-fidelity reconstruction in the pixel space. We therefore perform pixel-space fine-tuning on the decoded output of the one-step prediction.

Starting from $\mathbf{z}_{\mathrm{LQ}}$ at $t=\tau$, the restored latent is computed using the same update as that employed during inference:
\begin{equation}
  \widehat{\mathbf{z}}
  =
  \mathbf{z}_{\mathrm{LQ}}
  -
  \sigma_\tau
  v_\theta(
    \mathbf{z}_{\mathrm{LQ}},
    \tau,
    \mathbf{c}_{\varnothing}).
  \label{eq:one_step}
\end{equation}
The restored video is then obtained using the frozen Wan VAE decoder:
\begin{equation}
  \widehat{\mathbf{x}}
  =
  \mathcal{G}_{\mathrm{Wan}}(\widehat{\mathbf{z}}).
  \label{eq:wan_decoding}
\end{equation}

We supervise the decoded output using an equally weighted combination of the pixel-wise $\ell_1$ loss and the DISTS perceptual loss~\cite{ding2020dists}:
\begin{equation}
  \mathcal{L}_{\mathrm{II}}
  =
  \mathcal{L}_{\ell_1}
  (\widehat{\mathbf{x}},\mathbf{x}_{\mathrm{HQ}})
  +
  \mathcal{L}_{\mathrm{DISTS}}
  (\widehat{\mathbf{x}},\mathbf{x}_{\mathrm{HQ}}).
  \label{eq:pixel_loss}
\end{equation}
The Wan VAE decoder remains frozen, while gradients are propagated through it to update the DiT parameters $\theta$. Stage~II therefore aligns the latent prediction with the pixel-space reconstruction objective without modifying the inference architecture.

\section{Experiments}
\label{experiments}
In this section, we evaluate the proposed FastVR on synthetic and real-world benchmarks and compare it with state-of-the-art methods.

\subsection{Implementation Details}
\paragraph{Training dataset.}
To ensure high visual quality, we construct a training set of approximately 80,000 videos filtered based on resolution, scene transitions, and no-reference quality scores. During training, we resize each video so that its shorter side is 1024 pixels, then center-crop it to $1024 \times 1024$ pixels. In both training stages, we sample 45-frame clips with randomly selected temporal strides and use the RealBasicVSR degradation pipeline to synthesize the corresponding low-quality clips on the fly.

\paragraph{Optimization.}
We initialize the DiT from Wan2.2-TI2V-5B and use empty text prompts during training. We use the AdamW optimizer with $(\beta_1,\beta_2)=(0.9,0.95)$ and train in bfloat16 precision on 32 NVIDIA H100-80G GPUs, with a global batch size of 32. The learning rates are $2\times10^{-5}$ and $1\times10^{-5}$ for the first and second stages, respectively. We train for approximately 10,000 iterations in total across the two stages, using learning rate warm-up followed by cosine decay. We also use gradient checkpointing and apply gradient clipping with a threshold of 1.

\paragraph{Inference.}
We perform a single step from the point in the shifted schedule closest to timestep 399. We run the DiT with chunk-wise causal attention along the temporal dimension, using a chunk size of 3 latent frames. For spatial tiling, we use overlapping $64\times64$ latent tiles. The latency includes the execution time of all enabled modules but excludes file I/O and model loading.

\subsection{Evaluation and Metrics.}
We compare FastVR with state-of-the-art video restoration methods, including STAR \cite{star}, SeedVR-7B \cite{seedvr}, Vivid-VR \cite{vivid}, DOVE \cite{dove}, SATB-VR (one-step) \cite{satbvr}, SeedVR2-7B \cite{seedvr2}, and the Tiny and Full variants of FlashVSR \cite{flashvsr}. We use the same benchmarks as Vivid-VR, covering both synthetic datasets (SPMCS, UDM10, and YouHQ40) and real-world datasets (VideoLQ and UGC50). For real-world videos without ground-truth references, we use no-reference image quality metrics (NIQE, MANIQA, MUSIQ, and CLIP-IQA) and the video quality metric DOVER. For the synthetic benchmarks, we additionally report the full-reference metrics PSNR, SSIM, and LPIPS.

\subsection{Quantitative Results}

Table~\ref{tab:quantitative-results} presents comparisons on five synthetic and real-world benchmarks. FastVR achieves the highest MUSIQ and CLIP-IQA scores on all five datasets and ranks first or second in DOVER. It also obtains the best results across all reported metrics on VideoLQ. On synthetic benchmarks, FastVR improves LPIPS over both FlashVSR variants, although DOVE and SeedVR2 generally retain advantages in full-reference metrics. Overall, these results demonstrate that FastVR delivers strong no-reference restoration quality across diverse degradations with a single diffusion step.

\begin{table*}[!t]
\centering
\renewcommand\arraystretch{1.1}
\setlength{\tabcolsep}{3pt}
\footnotesize
\begin{tabular}{c|c|ccc|ccccc|c}
\toprule
Datasets & Metrics & \makecell{STAR} & \makecell{SeedVR} & \makecell{Vivid-VR} & \makecell{DOVE} & \makecell{SATB-VR} & \makecell{SeedVR2} & \makecell{FlashVSR-\\Tiny} & \makecell{FlashVSR-\\Full} & \makecell{Ours} \\ \hline
\multirow{8}{*}{SPMCS} & PSNR $\uparrow$ & 24.18 & 24.08 & 21.73 & {\color[HTML]{3166FF} \underline{24.80}} & 24.18 & {\color[HTML]{F94848} \textbf{26.07}} & 23.57 & 23.44 & 23.50 \\
 & SSIM $\uparrow$ & 0.720 & 0.689 & 0.604 & {\color[HTML]{3166FF} \underline{0.754}} & 0.707 & {\color[HTML]{F94848} \textbf{0.777}} & 0.675 & 0.670 & 0.662 \\
 & LPIPS $\downarrow$ & 0.301 & 0.263 & 0.278 & {\color[HTML]{F94848} \textbf{0.168}} & 0.197 & {\color[HTML]{3166FF} \underline{0.191}} & 0.223 & 0.226 & 0.221 \\ \cline{2-11}
 & NIQE $\downarrow$ & 7.058 & 4.514 & {\color[HTML]{3166FF} \underline{3.457}} & 4.031 & 4.047 & 4.969 & 3.496 & {\color[HTML]{F94848} \textbf{3.278}} & 3.505 \\
 & MANIQA $\uparrow$ & 0.229 & 0.315 & {\color[HTML]{F94848} \textbf{0.410}} & 0.346 & 0.384 & 0.305 & 0.361 & 0.381 & {\color[HTML]{3166FF} \underline{0.400}} \\
 & MUSIQ $\uparrow$ & 30.62 & 56.99 & {\color[HTML]{3166FF} \underline{70.03}} & 63.29 & 67.82 & 53.23 & 66.27 & 67.91 & {\color[HTML]{F94848} \textbf{71.23}} \\
 & CLIP-IQA $\uparrow$ & 0.254 & 0.347 & 0.483 & 0.410 & 0.514 & 0.325 & 0.512 & {\color[HTML]{3166FF} \underline{0.571}} & {\color[HTML]{F94848} \textbf{0.586}} \\
 & DOVER $\uparrow$ & 4.266 & 9.779 & {\color[HTML]{3166FF} \underline{11.35}} & 9.898 & 10.65 & 8.625 & 10.33 & 10.38 & {\color[HTML]{F94848} \textbf{11.71}} \\ \hline
\multirow{8}{*}{UDM10} & PSNR $\uparrow$ & 27.29 & 27.80 & 24.54 & {\color[HTML]{F94848} \textbf{30.53}} & 28.67 & {\color[HTML]{3166FF} \underline{29.04}} & 26.82 & 26.36 & 28.76 \\
 & SSIM $\uparrow$ & 0.855 & 0.848 & 0.761 & {\color[HTML]{F94848} \textbf{0.894}} & 0.859 & {\color[HTML]{3166FF} \underline{0.884}} & 0.806 & 0.797 & 0.842 \\
 & LPIPS $\downarrow$ & 0.167 & 0.148 & 0.243 & {\color[HTML]{F94848} \textbf{0.101}} & 0.150 & {\color[HTML]{3166FF} \underline{0.117}} & 0.172 & 0.182 & 0.154 \\ \cline{2-11}
 & NIQE $\downarrow$ & 6.072 & 5.345 & 4.046 & 5.055 & 4.283 & 5.641 & 3.941 & {\color[HTML]{3166FF} \underline{3.779}} & {\color[HTML]{F94848} \textbf{3.742}} \\
 & MANIQA $\uparrow$ & 0.260 & 0.264 & 0.359 & 0.296 & {\color[HTML]{F94848} \textbf{0.381}} & 0.262 & 0.341 & {\color[HTML]{3166FF} \underline{0.364}} & {\color[HTML]{F94848} \textbf{0.381}} \\
 & MUSIQ $\uparrow$ & 45.38 & 50.29 & 64.71 & 55.17 & {\color[HTML]{3166FF} \underline{65.83}} & 48.91 & 62.49 & 65.07 & {\color[HTML]{F94848} \textbf{67.50}} \\
 & CLIP-IQA $\uparrow$ & 0.289 & 0.273 & 0.426 & 0.340 & 0.507 & 0.272 & 0.494 & {\color[HTML]{3166FF} \underline{0.556}} & {\color[HTML]{F94848} \textbf{0.568}} \\
 & DOVER $\uparrow$ & 9.454 & 9.349 & {\color[HTML]{F94848} \textbf{11.97}} & 10.41 & 10.98 & 8.752 & 11.52 & 11.60 & {\color[HTML]{3166FF} \underline{11.86}} \\ \hline
\multirow{8}{*}{YouHQ40} & PSNR $\uparrow$ & 22.92 & 22.46 & 21.31 & {\color[HTML]{F94848} \textbf{24.10}} & 23.67 & {\color[HTML]{3166FF} \underline{24.00}} & 22.77 & 22.56 & 23.10 \\
 & SSIM $\uparrow$ & 0.657 & 0.621 & 0.579 & {\color[HTML]{3166FF} \underline{0.688}} & 0.657 & {\color[HTML]{F94848} \textbf{0.693}} & 0.608 & 0.602 & 0.621 \\
 & LPIPS $\downarrow$ & 0.433 & {\color[HTML]{3166FF} \underline{0.240}} & 0.357 & 0.283 & 0.281 & {\color[HTML]{F94848} \textbf{0.185}} & 0.300 & 0.290 & 0.271 \\ \cline{2-11}
 & NIQE $\downarrow$ & 6.744 & 4.243 & {\color[HTML]{3166FF} \underline{3.410}} & 4.456 & 4.004 & 4.576 & 3.603 & 3.465 & {\color[HTML]{F94848} \textbf{3.207}} \\
 & MANIQA $\uparrow$ & 0.240 & 0.315 & {\color[HTML]{3166FF} \underline{0.372}} & 0.304 & 0.354 & 0.314 & 0.347 & 0.367 & {\color[HTML]{F94848} \textbf{0.380}} \\
 & MUSIQ $\uparrow$ & 36.36 & 61.91 & {\color[HTML]{3166FF} \underline{70.55}} & 60.65 & 67.91 & 59.34 & 66.87 & 69.62 & {\color[HTML]{F94848} \textbf{72.91}} \\
 & CLIP-IQA $\uparrow$ & 0.279 & 0.360 & 0.447 & 0.356 & 0.486 & 0.336 & 0.527 & {\color[HTML]{3166FF} \underline{0.590}} & {\color[HTML]{F94848} \textbf{0.620}} \\
 & DOVER $\uparrow$ & 7.868 & 14.00 & {\color[HTML]{F94848} \textbf{14.61}} & 12.52 & 13.25 & 12.80 & 13.70 & 13.84 & {\color[HTML]{3166FF} \underline{14.57}} \\ \hline
\multirow{5}{*}{VideoLQ} & NIQE $\downarrow$ & 5.789 & 4.994 & 4.371 & 5.049 & 4.260 & 5.674 & 4.060 & {\color[HTML]{3166FF} \underline{3.892}} & {\color[HTML]{F94848} \textbf{3.759}} \\
 & MANIQA $\uparrow$ & 0.271 & 0.223 & 0.319 & 0.272 & {\color[HTML]{3166FF} \underline{0.356}} & 0.221 & 0.278 & 0.299 & {\color[HTML]{F94848} \textbf{0.361}} \\
 & MUSIQ $\uparrow$ & 50.52 & 46.49 & 62.47 & 55.11 & {\color[HTML]{3166FF} \underline{65.59}} & 43.41 & 57.54 & 61.88 & {\color[HTML]{F94848} \textbf{69.17}} \\
 & CLIP-IQA $\uparrow$ & 0.265 & 0.229 & 0.338 & 0.271 & {\color[HTML]{3166FF} \underline{0.436}} & 0.220 & 0.348 & 0.405 & {\color[HTML]{F94848} \textbf{0.446}} \\
 & DOVER $\uparrow$ & 8.758 & 7.240 & {\color[HTML]{3166FF} \underline{9.743}} & 8.780 & 9.577 & 6.331 & 8.954 & 9.360 & {\color[HTML]{F94848} \textbf{9.761}} \\ \hline
\multirow{5}{*}{UGC50} & NIQE $\downarrow$ & 5.754 & 5.662 & 4.361 & 5.493 & 4.672 & 6.230 & 4.083 & {\color[HTML]{F94848} \textbf{3.887}} & {\color[HTML]{3166FF} \underline{3.891}} \\
 & MANIQA $\uparrow$ & 0.325 & 0.262 & 0.376 & 0.320 & {\color[HTML]{F94848} \textbf{0.402}} & 0.253 & 0.354 & 0.372 & {\color[HTML]{3166FF} \underline{0.379}} \\
 & MUSIQ $\uparrow$ & 55.01 & 49.76 & 67.61 & 57.82 & {\color[HTML]{3166FF} \underline{68.52}} & 46.12 & 63.85 & 65.66 & {\color[HTML]{F94848} \textbf{68.92}} \\
 & CLIP-IQA $\uparrow$ & 0.353 & 0.305 & 0.450 & 0.353 & {\color[HTML]{3166FF} \underline{0.571}} & 0.276 & 0.516 & 0.563 & {\color[HTML]{F94848} \textbf{0.587}} \\
 & DOVER $\uparrow$ & 10.92 & 10.47 & {\color[HTML]{F94848} \textbf{14.46}} & 11.84 & 13.40 & 8.209 & 13.40 & 13.29 & {\color[HTML]{3166FF} \underline{13.86}} \\
\bottomrule
\end{tabular}
\vspace{-2mm}
\caption{Quantitative comparisons on benchmarks, including synthetic (SPMCS, UDM10, YouHQ40) and real-world (VideoLQ, UGC50) videos. The best and second-best results are marked in {\color[HTML]{F94848} \textbf{bold}} and {\color[HTML]{3166FF} \underline{underline}}, respectively. Ties at the displayed precision share the same rank.}
\vspace{-4mm}
\label{tab:quantitative-results}
\end{table*}

\subsection{Qualitative Results}

\begin{figure}[!t]
    \centering
    \includegraphics[width=0.87\linewidth, height=\textheight, keepaspectratio, pagebox=cropbox]{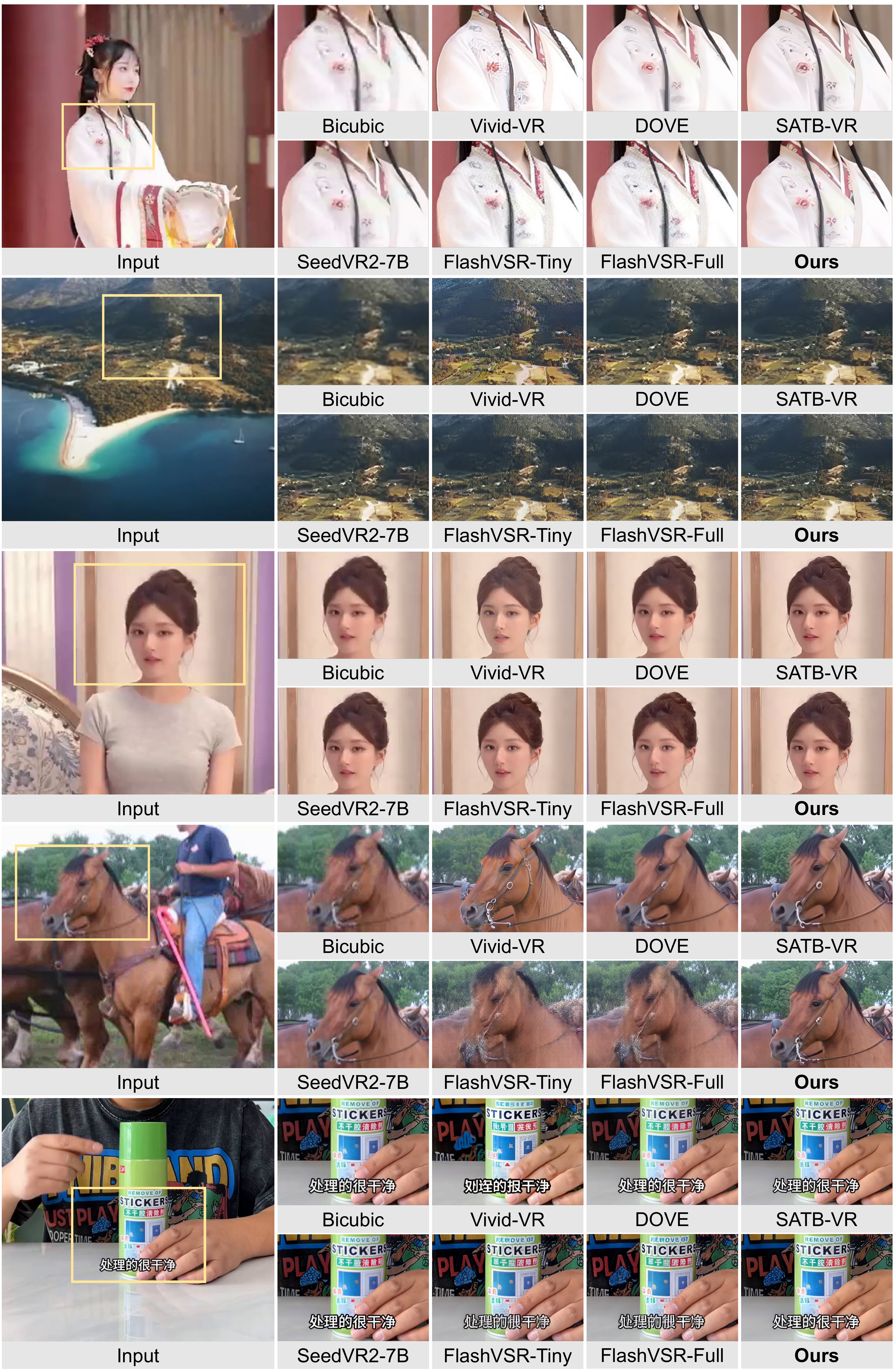}
    \caption{Qualitative comparisons on real-world videos, showing clothing, landscapes, portraits, animals, and text. The boxed regions in the input frames are enlarged for comparison. \textbf{Zoom in for details.}}
    \label{fig:qualitative-real-world}
\end{figure}

\begin{figure}[!t]
    \centering
    \includegraphics[width=0.87\linewidth, height=\textheight, keepaspectratio, pagebox=cropbox]{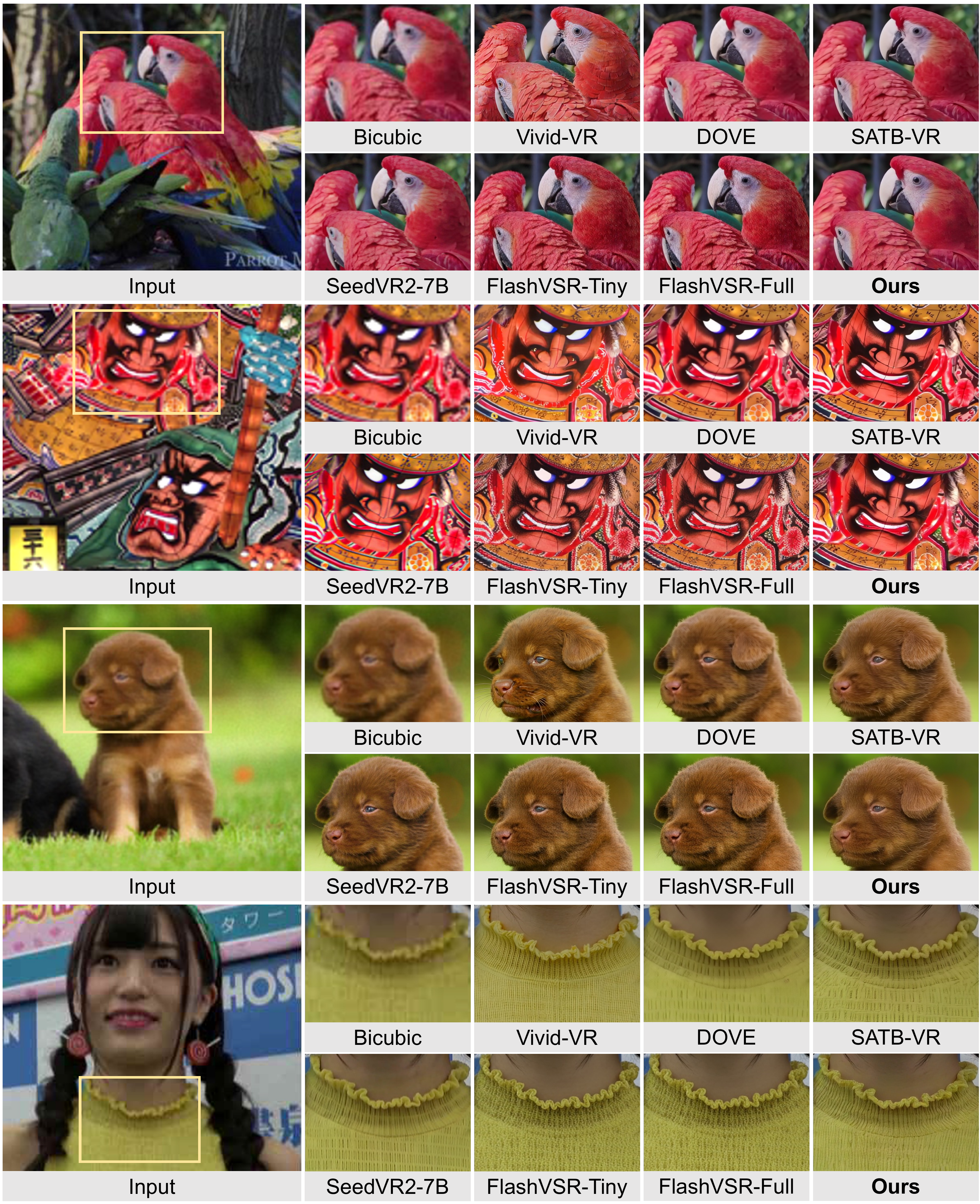}
    \caption{Qualitative comparisons on synthetically degraded videos, showing feathers, illustrated patterns, animal fur, and knitted fabric. The boxed regions in the input frames are enlarged for comparison. \textbf{Zoom in for details.}}
    \label{fig:qualitative-synthetic}
\end{figure}

Figures~\ref{fig:qualitative-real-world} and~\ref{fig:qualitative-synthetic} present visual comparisons on real-world and synthetic videos. 
On real-world footage, FastVR reconstructs clearer embroidery, hair strands, and horse-bridle details, while avoiding the pronounced texture artifacts visible in some competing results. It also preserves the subtitle characters more faithfully in the final example, where Vivid-VR and the FlashVSR variants introduce visible distortions.
In the synthetic examples, FastVR recovers fine fur and fabric detail while maintaining clear object contours. Compared with the FlashVSR variants, it produces less granular textures in the illustrated face and knitted garment. 
Overall, these examples suggest that FastVR achieves a favorable balance between detail recovery and artifact suppression using one-step diffusion inference.

\enlargethispage{\baselineskip}
\section{Conclusion}

We presented FastVR, a one-step diffusion framework for efficient streaming video restoration. By combining a lightweight VAE with chunk-wise causal attention and KV caching, FastVR reduces autoencoding and denoising overhead while keeping intermediate memory bounded during long-video inference. Its two-stage training combines continuous latent-space trajectory learning and velocity consistency regularization with pixel-space refinement. Experiments on synthetic and real-world benchmarks demonstrate strong no-reference quality and competitive visual detail, together with favorable inference speed and memory usage. These results highlight the potential of jointly designing efficient inference and restoration-oriented training for practical high-resolution video restoration.

\clearpage
{
\bibliographystyle{plain}
\bibliography{references}

@article{dove,
  title={{DOVE}: Efficient One-Step Diffusion Model for Real-World Video Super-Resolution},
  author={Chen, Zheng and Zou, Zichen and Zhang, Kewei and Su, Xiongfei and Yuan, Xin and Guo, Yong and Zhang, Yulun},
  journal={arXiv preprint arXiv:2505.16239},
  year={2025},
  eprint={2505.16239},
  archivePrefix={arXiv}
}

@article{seedvr2,
  title={{SeedVR2}: One-Step Video Restoration via Diffusion Adversarial Post-Training},
  author={Wang, Jianyi and Lin, Shanchuan and Lin, Zhijie and Ren, Yuxi and Wei, Meng and Yue, Zongsheng and Zhou, Shangchen and Chen, Hao and Zhao, Yang and Yang, Ceyuan and Xiao, Xuefeng and Loy, Chen Change and Jiang, Lu},
  journal={arXiv preprint arXiv:2506.05301},
  year={2025},
  eprint={2506.05301},
  archivePrefix={arXiv}
}

@article{vivid,
  title={{Vivid-VR}: Distilling Concepts from Text-to-Video Diffusion Transformer for Photorealistic Video Restoration},
  author={Bai, Haoran and Chen, Xiaoxu and Yang, Canqian and He, Zongyao and Deng, Sibin and Chen, Ying},
  journal={arXiv preprint arXiv:2508.14483},
  year={2025},
  eprint={2508.14483},
  archivePrefix={arXiv}
}

@article{flashvsr,
  title={{FlashVSR}: Towards Real-time Diffusion-Based Streaming Video Super-Resolution},
  author={Zhuang, Junhao and Guo, Shi and Cai, Xin and Li, Xiaohui and Liu, Yihao and Yuan, Chun and Xue, Tianfan},
  journal={arXiv preprint arXiv:2510.12747},
  year={2025},
  eprint={2510.12747},
  archivePrefix={arXiv}
}

@article{duovsr,
  title={{DUO-VSR}: Dual-Stream Distillation for One-Step Video Super-Resolution},
  author={Lv, Zhengyao and Xia, Menghan and Wang, Xintao and Wong, Kwan-Yee K},
  journal={arXiv preprint arXiv:2603.22271},
  year={2026},
  eprint={2603.22271},
  archivePrefix={arXiv}
}

@article{satbvr,
  title={{SATB-VR}: Training Few-Step Video Restoration Diffusion Model using SNR-Aware Trajectory Blending},
  author={Bai, Haoran and Chen, Xiaoxu and Liu, Xiaoyu and Yue, Zongsheng and Deng, Sibin and Zuo, Wangmeng and Chen, Ying},
  journal={arXiv preprint arXiv:2606.28677},
  year={2026},
  eprint={2606.28677},
  archivePrefix={arXiv}
}

@article{wan,
  title={Wan: Open and Advanced Large-Scale Video Generative Models},
  author={{Wan Team} and Wang, Ang and Ai, Baole and Wen, Bin and Mao, Chaojie and Xie, Chen-Wei and Chen, Di and Yu, Feiwu and Zhao, Haiming and Yang, Jianxiao and others},
  journal={arXiv preprint arXiv:2503.20314},
  year={2025},
  eprint={2503.20314},
  archivePrefix={arXiv}
}

@inproceedings{basicvsr,
  title={{BasicVSR}: The Search for Essential Components in Video Super-Resolution and Beyond},
  author={Chan, Kelvin C. K. and Wang, Xintao and Yu, Ke and Dong, Chao and Loy, Chen Change},
  booktitle={Proceedings of the IEEE/CVF Conference on Computer Vision and Pattern Recognition},
  pages={4947--4956},
  year={2021}
}

@inproceedings{basicvsrpp,
  title={{BasicVSR++}: Improving Video Super-Resolution with Enhanced Propagation and Alignment},
  author={Chan, Kelvin C. K. and Zhou, Shangchen and Xu, Xiangyu and Loy, Chen Change},
  booktitle={Proceedings of the IEEE/CVF Conference on Computer Vision and Pattern Recognition},
  pages={5972--5981},
  year={2022}
}

@inproceedings{upscale,
  title={{Upscale-A-Video}: Temporal-Consistent Diffusion Model for Real-World Video Super-Resolution},
  author={Zhou, Shangchen and Yang, Peiqing and Wang, Jianyi and Luo, Yihang and Loy, Chen Change},
  booktitle={Proceedings of the IEEE/CVF Conference on Computer Vision and Pattern Recognition},
  pages={2535--2545},
  year={2024}
}

@inproceedings{mgld-vsr,
  title={Motion-guided latent diffusion for temporally consistent real-world video super-resolution},
  author={Yang, Xi and He, Chenhang and Ma, Jianqi and Zhang, Lei},
  booktitle={European conference on computer vision},
  pages={224--242},
  year={2024},
  organization={Springer}
}

@inproceedings{diffvsr,
  title={Diffvsr: Revealing an effective recipe for taming robust video super-resolution against complex degradations},
  author={Li, Xiaohui and Liu, Yihao and Cao, Shuo and Chen, Ziyan and Zhuang, Shaobin and Chen, Xiangyu and He, Yinan and Wang, Yi and Qiao, Yu},
  booktitle={2025 IEEE/CVF International Conference on Computer Vision (ICCV)},
  pages={15319--15328},
  year={2025},
  organization={IEEE}
}

@inproceedings{cogvideox,
  title={Cogvideox: Text-to-video diffusion models with an expert transformer},
  author={Yang, Zhuoyi and Teng, Jiayan and Zheng, Wendi and Ding, Ming and Huang, Shiyu and Xu, Jiazheng and Yang, Yuanming and Hong, Wenyi and Zhang, Xiaohan and Feng, Guanyu and others},
  booktitle={International Conference on Learning Representations},
  volume={2025},
  pages={83048--83077},
  year={2025}
}

@article{ding2020dists,
  title={Image Quality Assessment: Unifying Structure and Texture Similarity},
  author={Ding, Keyan and Ma, Kede and Wang, Shiqi and Simoncelli, Eero P.},
  journal={IEEE Transactions on Pattern Analysis and Machine Intelligence},
  volume={44},
  number={5},
  pages={2567--2581},
  year={2020},
  publisher={IEEE}
}

@inproceedings{zhang2018lpips,
  title={The Unreasonable Effectiveness of Deep Features as a Perceptual Metric},
  author={Zhang, Richard and Isola, Phillip and Efros, Alexei A. and Shechtman, Eli and Wang, Oliver},
  booktitle={Proceedings of the IEEE/CVF Conference on Computer Vision and Pattern Recognition},
  pages={586--595},
  year={2018}
}

@inproceedings{star,
  title={Star: Spatial-temporal augmentation with text-to-video models for real-world video super-resolution},
  author={Xie, Rui and Liu, Yinhong and Zhou, Penghao and Zhao, Chen and Zhou, Jun and Zhang, Kai and Zhang, Zhenyu and Yang, Jian and Yang, Zhenheng and Tai, Ying},
  booktitle={2025 IEEE/CVF International Conference on Computer Vision (ICCV)},
  pages={17108--17118},
  year={2025},
  organization={IEEE}
}

@misc{tae,
  author = {Boer Bohan, Ollin},
  title = {TAEHV: Tiny AutoEncoder for Hunyuan Video},
  year = {2025},
  howpublished = {\url{https://github.com/madebyollin/taehv}},
}

@article{ddpm,
  title={Denoising diffusion probabilistic models},
  author={Ho, Jonathan and Jain, Ajay and Abbeel, Pieter},
  journal={Advances in neural information processing systems},
  volume={33},
  pages={6840--6851},
  year={2020}
}

@article{score-model,
  title={Score-based generative modeling through stochastic differential equations},
  author={Song, Yang and Sohl-Dickstein, Jascha and Kingma, Diederik P and Kumar, Abhishek and Ermon, Stefano and Poole, Ben},
  journal={arXiv preprint arXiv:2011.13456},
  year={2020}
}

@article{isr1,
  title={Exploiting diffusion prior for real-world image super-resolution},
  author={Wang, Jianyi and Yue, Zongsheng and Zhou, Shangchen and Chan, Kelvin CK and Loy, Chen Change},
  journal={International Journal of Computer Vision},
  volume={132},
  number={12},
  pages={5929--5949},
  year={2024},
  publisher={Springer}
}

@inproceedings{isr2,
  title={Scaling up to excellence: Practicing model scaling for photo-realistic image restoration in the wild},
  author={Yu, Fanghua and Gu, Jinjin and Li, Zheyuan and Hu, Jinfan and Kong, Xiangtao and Wang, Xintao and He, Jingwen and Qiao, Yu and Dong, Chao},
  booktitle={Proceedings of the IEEE/CVF Conference on Computer Vision and Pattern Recognition},
  pages={25669--25680},
  year={2024},
  organization={IEEE}
}

@inproceedings{isr3,
  title={Arbitrary-steps image super-resolution via diffusion inversion},
  author={Yue, Zongsheng and Liao, Kang and Loy, Chen Change},
  booktitle={Proceedings of the IEEE/CVF Conference on Computer Vision and Pattern Recognition},
  pages={23153--23163},
  year={2025},
  organization={IEEE}
}

@inproceedings{seedvr,
   title={SeedVR: Seeding Infinity in Diffusion Transformer Towards Generic Video Restoration},
   author={Wang, Jianyi and Lin, Zhijie and Wei, Meng and Zhao, Yang and Yang, Ceyuan and Loy, Chen Change and Jiang, Lu},
   booktitle={Proceedings of the IEEE/CVF Conference on Computer Vision and Pattern Recognition},
   year={2025}
}

@inproceedings{dmd,
  title={One-step diffusion with distribution matching distillation},
  author={Yin, Tianwei and Gharbi, Micha{\"e}l and Zhang, Richard and Shechtman, Eli and Durand, Fredo and Freeman, William T and Park, Taesung},
  booktitle={Proceedings of the IEEE/CVF Conference on Computer Vision and Pattern Recognition},
  pages={6613--6623},
  year={2024},
  organization={IEEE}
}

@inproceedings{dmd2,
  title={Improved distribution matching distillation for fast image synthesis},
  author={Yin, Tianwei and Gharbi, Micha{\"e}l and Park, Taesung and Zhang, Richard and Shechtman, Eli and Durand, Fredo and Freeman, William T},
  booktitle={Advances in neural information processing systems},
  volume={37},
  pages={47455--47487},
  year={2024}
}

@inproceedings{apt,
  title={Diffusion Adversarial Post-Training for One-Step Video Generation},
  author={Lin, Shanchuan and Xia, Xin and Ren, Yuxi and Yang, Ceyuan and Xiao, Xuefeng and Jiang, Lu},
  booktitle={Proceedings of the 42nd International Conference on Machine Learning},
  pages={37959--37974},
  year={2025},
  volume={267},
  publisher={PMLR},
}

@inproceedings{liu2023flow,
    title={Flow Straight and Fast: Learning to Generate and Transfer Data with Rectified Flow},
    author={Xingchao Liu and Chengyue Gong and qiang liu},
    booktitle={The Eleventh International Conference on Learning Representations},
    year={2023},
}

@inproceedings{realbasicvsr,
  author = {Chan, Kelvin C.K. and Zhou, Shangchen and Xu, Xiangyu and Loy, Chen Change},
  title = {Investigating Tradeoffs in Real-World Video Super-Resolution},
  booktitle = {Proceedings of the IEEE/CVF Conference on Computer Vision and Pattern Recognition},
  year = {2022}
}
}

\end{document}